\documentclass[review,12pt]{elsarticle}

\usepackage[misc]{ifsym}
\usepackage[colorinlistoftodos]{todonotes}

\usepackage[caption=false]{subfig}
\usepackage{float}

\usepackage{multirow}
\usepackage{rotating}

\usepackage{hyperref}

\usepackage{mathtools,etoolbox}
\DeclarePairedDelimiterX{\abs}[1]{\lvert}{\rvert}{\ifblank{#1}{{}\cdot{}}{#1}}

\usepackage{graphicx}
\usepackage{amssymb}

\usepackage{manyfoot}

\DeclareNewFootnote{A}
\DeclareNewFootnote{B}

\let\footnoteSym\footnoteA

\newcommand{\FF}{\mathbf{F}}
\newcommand{\XX}{\mathbf{X}}
\newcommand{\RR}{\mathbb{R}}
\newcommand{\UU}{\mathbf{U}}

\newcommand{\WW}{\mathbf{W}}
\newcommand{\uu}{\mathbf{u}}

\newcommand{\zz}{\mathbf{z}}
\newcommand{\xx}{\mathbf{x}}

\makeatletter
\def\Hline{
\noalign{\ifnum0=`}\fi\hrule \@height 1pt \futurelet
\reserved@a\@xhline}
\makeatother

\journal{Neurocomputing}

\begin{document}

\begin{frontmatter}


\title{USE-Net: incorporating Squeeze-and-Excitation blocks into U-Net for prostate zonal segmentation \\of multi-institutional MRI datasets}



\author[Unimib,IBFM,CamRadiol,CRUK]{Leonardo Rundo\corref{corr}\footnoteSym[2]{These authors contributed equally.}\textsuperscript{,}}
\author[Todai]{Changhee Han$^{\dagger,}$}
\author[Todai]{Yudai Nagano}
\author[Todai]{Jin Zhang}
\author[Todai]{Ryuichiro Hataya}
\author[IBFM]{Carmelo Militello}
\author[Unimib,CamHaem,Sanger]{Andrea Tangherloni}
\author[Unimib,SYSBIO]{\\Marco S. Nobile}
\author[Unimib]{Claudio Ferretti}
\author[Unimib]{Daniela Besozzi}
\author[IBFM]{\\Maria Carla Gilardi}
\author[Unipa]{Salvatore Vitabile}
\author[Unimib,SYSBIO]{Giancarlo Mauri}
\author[Todai]{\\Hideki Nakayama}
\author[Unibg,SYSBIO]{Paolo Cazzaniga}

\cortext[corr]{Corresponding author.\\ \textit{E-mail address}: leonardo.rundo@disco.unimib.it; lr495@cam.ac.uk (L. Rundo)}

\address[Unimib]{Department of Informatics, Systems and Communication, University of Milano-Bicocca, 20126 Milan, Italy}
\address[IBFM]{Institute of Molecular Bioimaging and Physiology, Italian National Research Council, 90015 Cefal\'u (PA), Italy}
\address[CamRadiol]{Department of Radiology, University of Cambridge, CB2 0QQ Cambridge, UK}
\address[CRUK]{Cancer Research UK Cambridge Centre, CB2 0RE Cambridge, UK}
\address[Todai]{Graduate School of Information Science and Technology, The University of Tokyo, 113-8656 Tokyo, Japan}
\address[CamHaem]{Department of Haematology, University of Cambridge, CB2 0XY Cambridge, UK}
\address[Sanger]{Wellcome Trust Sanger Institute, Wellcome Trust Genome Campus,\\CB10 1SA Hinxton, UK}
\address[SYSBIO]{SYSBIO.IT Centre of Systems Biology, 20126 Milan, Italy}
\address[Unipa]{Department of Biomedicine, Neuroscience and Advanced Diagnostics,\\University of Palermo, 90127 Palermo, Italy}
\address[Unibg]{Department of Human and Social Sciences, University of Bergamo,\\24129 Bergamo, Italy}

\begin{abstract}
Prostate cancer is the most common malignant tumors in men but prostate Magnetic Resonance Imaging (MRI) analysis remains challenging.
Besides whole prostate gland segmentation, the capability to differentiate between the blurry boundary of the Central Gland (CG) and Peripheral Zone (PZ) can lead to differential diagnosis, since the frequency and severity of tumors differ in these regions.
To tackle the prostate zonal segmentation task, we propose a novel Convolutional Neural Network (CNN), called USE-Net, which incorporates Squeeze-and-Excitation (SE) blocks into U-Net, i.e., one of the most effective CNNs in biomedical image segmentation.
Especially, the SE blocks are added after every Encoder (Enc USE-Net) or Encoder-Decoder block (Enc-Dec USE-Net).
This study evaluates the generalization ability of CNN-based architectures on three T2-weighted MRI datasets, each one consisting of a different number of patients and heterogeneous image characteristics, collected by different institutions.
The following mixed scheme is used for training/testing: (\textit{i}) training on either each individual dataset or multiple prostate MRI datasets and (\textit{ii}) testing on all three datasets with all possible training/testing combinations.
USE-Net is compared against three state-of-the-art CNN-based architectures (i.e., U-Net, pix2pix, and Mixed-Scale Dense Network), along with a semi-automatic continuous max-flow model.
The results show that training on the union of the datasets generally outperforms training on each dataset separately, allowing for both intra-/cross-dataset generalization.
Enc USE-Net shows good overall generalization under any training condition, while Enc-Dec USE-Net remarkably outperforms the other methods when trained on all datasets.
These findings reveal that the SE blocks' adaptive feature recalibration provides excellent cross-dataset generalization when testing is performed on samples of the datasets used during training.
Therefore, we should consider multi-dataset training and SE blocks together as mutually indispensable methods to draw out each other's full potential.
In conclusion, adaptive mechanisms (e.g., feature recalibration) may be a valuable solution in medical imaging applications involving multi-institutional settings.
\end{abstract}


\end{frontmatter}


\section{Introduction}
\label{sec:Intro}
According to the American Cancer Society, in 2019 the Prostate Cancer (PCa) is expected to be the most common malignant tumor with the second highest mortality for American males~\cite{siegel2019}.
Given a clinical context, several imaging modalities can be used for PCa diagnosis, such as Transrectal Ultrasound (TRUS), Computed Tomography (CT), and Magnetic Resonance Imaging (MRI).
For an in-depth investigation, structural T1-weighted (T1w) and T2-weighted (T2w) MRI sequences can be combined with the functional information from Dynamic Contrast Enhanced MRI (DCE-MRI), Diffusion Weighted Imaging (DWI), and Magnetic Resonance Spectroscopic Imaging (MRSI)~\cite{lemaitre2015}.
Recent advancements in MRI scanners, especially those related to magnetic field strengths higher than $1.5$T, did not decrease the effect of magnetic susceptibility artifacts on prostate MR images, even though the shift from $1.5$T to $3$T theoretically leads to a doubled Signal-to-Noise Ratio (SNR)~\cite{rouviere2006}.
However, $3$T MRI scanners permitted to obtain high-quality images with less invasive procedures compared with $1.5$T, thanks to a pelvic coil that reduces prostate gland compression/deformation~\cite{kim2008,heijmink2007}.

Therefore, MRI plays a decisive role in PCa diagnosis and disease monitoring (even in an advanced status~\cite{padhani2017}), revealing the internal prostatic anatomy, prostatic margins, and PCa extent~\cite{villeirs2007}.
According to the zonal compartment system proposed by McNeal, the prostate Whole Gland (WG) can be partitioned into the Central Gland (CG) and Peripheral Zone (PZ) \cite{selman2011}.
In prostate imaging, T2w MRI serves as the principal sequence~\cite{scheenen2015}, thanks to its high resolution that enables to differentiate the hyper-intense PZ and hypo-intense CG in young male subjects~\cite{hoeks2011}.

Besides manual detection/delineation of the WG and PCa on MR images, distinguishing between the CG and PZ is clinically essential, since the frequency and severity of tumors differ in these regions~\cite{choi2007,niaf2012}.
As a matter of fact, the PZ harbors $70$-$80\%$ of PCa and represents a target for prostate biopsy~\cite{haffner2009}. Furthermore, the PZ volume ratio (i.e., the PZ volume divided by the WG volume) can be considered for PCa diagnostic refinement~\cite{chang2017}, while the CG volume ratio can help monitoring prostate hyperplasia~\cite{kirby2002}. 
Therefore, according to the Prostate Imaging-Reporting and Data System version 2 (PI-RADS\textsuperscript{TM} v2)~\cite{weinreb2016}, radiologists must perform a zonal partitioning before assessing the suspicion of PCa on multi-parametric MRI.
However, an improved PCa diagnosis requires a reliable and automatic zonal segmentation method, since  manual delineation is time-consuming and operator-dependent~\cite{rundo2017Inf,muller2015}.
Moreover, in clinical practice, the generalization ability among multi-institutional prostate MRI datasets is essential due to large anatomical inter-subject variability and the lack of a standardized pixel intensity representation for MRI (such as for CT-based radiodensity measurements expressed in Hounsfield units)~\cite{klein2008}.
Hence, we aim at automatically segmenting the prostate zones on three multi-institutional T2w MRI datasets to evaluate the generalization ability of Convolutional Neural Network (CNN)-based architectures.
This task is challenging because images from multi-institutional datasets are characterized by different contrasts, visual consistencies, and heterogeneous characteristics~\cite{vanOpbroek2015}.

In this work, we propose a novel CNN, called USE-Net, which incorporates Squeeze-and-Excitation (SE) blocks~\cite{hu2017} into U-Net after every Encoder (Enc USE-Net) or Encoder-Decoder block (Enc-Dec USE-Net).
The rationale behind the design of USE-Net is to exploit adaptive channel-wise feature recalibration to boost the generalization performance. The proposed USE-Net is conceived to outperform the state-of-the-art CNN-based architectures for segmentation in multi-institutional studies, whilst the SE blocks (initially proposed in~\cite{hu2017}) were originally designed to boost the performance only for classification and object detection \textit{via} feature recalibration, by capturing single dataset characteristics.
Unlike the original SE blocks placed in InceptionNet~\cite{szegedy2016} and ResNet~\cite{he2016} architectures, we introduced them into U-Net after the encoders and decoders to boost the segmentation performance with increased generalization ability, thanks to the representation of channel-wise relationships in multi-institutional clinical scenarios, analyzing multiple heterogeneous MRI datasets.
This study adopted a mixed scheme for cross- and intra-dataset generalization: (\textit{i}) training on either each individual dataset or multiple datasets, and (\textit{ii}) testing on all three datasets with all possible training/testing combinations.
To the best of our knowledge, this is the first CNN-based prostate zonal segmentation on T2w MRI alone.
By relying on both spatial overlap-/distance-based metrics, we compared USE-Net against three CNN-based architectures: U-Net, pix2pix, and Mixed-Scale Dense Network (MS-D Net)~\cite{pelt2017}, along with a semi-automatic continuous max-flow model~\cite{qiu2014}.

\paragraph{Contributions}
Our main contributions are:

\begin{itemize}
\item \textbf{Prostate zonal segmentation:}
our novel Enc-Dec USE-Net achieves accurate CG and PZ segmentation results on T2w MR images, remarkably outperforming the other competitor methods when trained on all datasets used for testing in multi-institutional scenarios.
\item \textbf{Cross-dataset generalization:} this first cross-dataset study, investigating all possible training/testing conditions among three different medical imaging datasets, shows that training on the union of multiple datasets generally outperforms training on each dataset during testing, realizing both intra-/cross-dataset generalization---thus, we may train CNNs by feeding samples from multiple different datasets for improving the performance.

\item \textbf{Deep Learning for medical imaging:} this research reveals that SE blocks provide excellent intra-dataset generalization in multi-insti\-tutional scenarios, when testing is performed on samples from the datasets used during training.
Therefore, adaptive mechanisms (e.g., feature recalibration in CNNs) may be a valuable solution in medical imaging applications involving multi-institutional settings.
\end{itemize}

The manuscript is structured as follows.
Section~\ref{sec:Background} outlines the background of prostate MRI zonal segmentation, especially related work on CNNs.
Section~\ref{sec:MatMet} describes the analyzed multi-institutional MRI datasets, the proposed USE-Net architectures, the investigated state-of-the-art CNN- and max-flow-based segmentation approaches, as well as the employed evaluation metrics; the experimental results are presented and discussed in Section~\ref{sec:Results}.
Finally, conclusive remarks and future directions of this work are given in Section~\ref{sec:Conclusions}.

\section{Related Work}
\label{sec:Background}
Due to the crucial role of MR image analysis in PCa diagnosis and staging~\cite{lemaitre2015}, researchers have paid specific attention to automatic WG detection/segmentation.
Classic methods mainly leveraged atlases~\cite{klein2008,martin2008} or statistical shape priors~\cite{martin2010}:
atlas-based approaches realized accurate segmentation when new prostate instances resemble the atlas, relying on a non-rigid registration algorithm~\cite{martin2010,toth2013}.
Unsupervised clustering techniques allowed for segmentation without manual labeling of large-scale MRI datasets~\cite{rundo2017Inf,rundo2018SIST}.
In the latest years, Deep Learning techniques \cite{litjens2017} have achieved accurate prostate segmentation results by using deep feature learning combined with shape models~\cite{guo2017} or location-prior maps~\cite{sun2017}.
Moreover, CNNs were used with patch-based ensemble learning~\cite{jia2018} or dense prediction schemes~\cite{milletari2016}.
In addition, end-to-end deep neural networks achieved outstanding results in automated PCa detection in multi-parametric MRI \cite{yang2017,wang2018}.

Differently from WG segmentation 
and PCa detection, less attention has been paid to CG and PZ segmentation despite its clinical importance in PCa diagnosis~\cite{niaf2012}.
In this context, classic Computer Vision techniques have been mainly exploited on T2w MRI.
For instance, early studies combined classifiers with statistical shape models~\cite{allen2006} or deformable models~\cite{yin2012}; Toth \textit{et al.}~\cite{toth2013} employed active appearance models with multiple level sets for simultaneous zonal segmentation; Qiu~\textit{et al.}~\cite{qiu2014} used a continuous max-flow model---the dual formulation of convex relaxed optimization with region consistency constraints~\cite{yuan2010}; in contrast, Makni~\textit{et al.}~\cite{makni2011} fused and processed 3D T2w, DWI, and contrast-enhanced T1w MR images by means of an evidential C-means algorithm~\cite{masson2008}.
As the first CNN-based method, Clark \textit{et al.}~\cite{clark2017} detected DWI MR images with prostate relying on Visual Geometry Group (VGG) net~\cite{simonyan2015}, and then sequentially segmented WG and CG using U-Net~\cite{ronneberger2015}.

Regarding the most recent computational methods in medical image segmentation, along with traditional Pattern Recognition techniques~\cite{rundo2018next}, significant advances have been proposed in CNN-based architectures.
For instance, to overcome the limitations related to accurate image annotations, DeepCut~\cite{rajchl2017} relies on weak bounding box labeling~\cite{rundo2017NC}.
This method aims at learning features for a CNN-based classifier from bounding box annotations.
Among the architectures devised for biomedical image segmentation~\cite{havaei2017,kamnitas2017}, U-Net~\cite{ronneberger2015} showed to be a noticeably successful solution, thanks to the combination of a contracting (i.e., encoding) path, for coarse-grained context detection, and a symmetric expanding (i.e., decoding) path, for fine-grained localization.
This fully CNN is capable of stable training with reduced samples.
The authors of V-Net~\cite{milletari2016} extended U-Net for volumetric medical image segmentation, by introducing also a different loss function based on the Dice Similarity Coefficient (\textit{DSC}).
Schlemper \textit{et al.}~\cite{schlemper2019} presented an Attention Gate (AG) model for medical imaging, which aims at focusing on target structures or organs.
AGs were introduced into the standard U-Net, so defining Attention U-Net, which achieved high performance in multi-class image segmentation without relying on multi-stage cascaded CNNs.
Recently MS-D Net~\cite{pelt2017} was shown to yield better segmentation results in biomedical images than U-Net~\cite{ronneberger2015} and SegNet~\cite{badrinarayanan2017}, by creating dense connections among features at different scales obtained by means of dilated convolutions.
By so doing, features at different scales can be contextually extracted using fewer parameters than full CNNs.
Finally, also image-to-image translation approaches---e.g., pix2pix~\cite{isola2016} that leverages conditional adversarial neural networks---were exploited for image segmentation.

However, no literature method so far coped with the generalization ability among multi-institutional MRI datasets, making their clinical applicability difficult~\cite{albadawy2018}.
In a previous work~\cite{rundoWIRN2018}, we compared existing CNN-based architectures---namely, SegNet~\cite{badrinarayanan2017}, U-Net~\cite{ronneberger2015}, and pix2pix~\cite{isola2016}---on two multi-institutional MRI datasets.
According to our results, U-Net generally achieves the most accurate performance.
Here, we thoroughly verify the intra-/cross-dataset generalization on three datasets from three different institutions, also proposing a novel architecture based on U-Net~\cite{ronneberger2015} incorporating SE blocks~\cite{hu2017}.
To the best of our knowledge, this is the first study on CNN-based prostate zonal segmentation on T2w MRI alone.

\section{Materials and Methods}
\label{sec:MatMet}

This section first describes the analyzed multi-institutional MRI datasets collected by different institutions.
Afterwards, we explain the proposed USE-Net, the other investigated CNN-based architectures, as well as a state-of-the-art prostate zonal segmentation method based on a continuous max-flow model~\cite{qiu2014}.
Finally, the used spatial overlap- and distance-based evaluation metrics are reported.

\subsection{Multi-institutional MRI Datasets}
\label{sec:Datasets}
We segment the CG and PZ from the WG on three completely different multi-parametric prostate MRI datasets, namely:
\begin{itemize}
\item[$\#1$] dataset ($21$ patients/$193$ MR slices with prostate), acquired with a whole body Philips Achieva 3T MRI scanner at the Cannizzaro Hospital (Catania, Italy) \cite{rundo2017Inf}.
MRI parameters: matrix size $= 288 \times 288$ pixels; slice thickness $= 3.0$ mm; inter-slice spacing $=4$ mm; pixel spacing $= 0.625$ mm; number of slices per image series (including slices without prostate) $= 18$.
Average patient age: $65.57 \pm 6.42$ years;
\item[$\#2$] Initiative for Collaborative Computer Vision Benchmarking (I2CVB) dataset ($19$ patients/$503$ MR slices with prostate), acquired with a whole body Siemens TIM 3T MRI scanner at the Hospital Center Regional University of Dijon-Bourgogne (Dijon, France) \cite{lemaitre2015}.
MRI parameters: matrix size $\in \{308 \times 384, 336 \times 448, 360 \times 448, 368 \times 448 \}$ pixels; slice thickness $= 1.25$ mm; inter-slice spacing $= 1.0$ mm; pixel spacing $\in \{0.676, 0.721, 0.881, 0.789 \}$ mm; number of slices per image series $= 64$.
Average patient age: $64.36 \pm 9.69$ years;
\item[$\#3$] National Cancer Institute -- International Symposium on Biomedical Imaging (NCI-ISBI) 2013 Automated Segmentation of Prostate Structures Challenge dataset ($40$ patients/$555$ MR slices with prostate) \textit{via} The Cancer Imaging Archive (TCIA) \cite{prior2017}, acquired with a whole body Siemens TIM 3T MRI scanner at Radboud University Medical Center (Nijmegen, The Netherlands) \cite{TCIA}.
The prostate structures were manually delineated by five experts.
MRI parameters: matrix size $\in \{256 \times 256, 320 \times 320, 384 \times 384\}$ pixels; slice thickness $\in \{3.0, 4.0\}$ mm; inter-slice spacing $\in \{3.6, 4.0\}$ mm; pixel spacing $\in \{0.500, 0.600, 0.625\}$ mm; number of slices per image series ranging from $15$ to $24$.
Average patient age: $63.90 \pm 7.17$ years.
\end{itemize}

All the analyzed MR images are encoded in the $16$-bit Digital Imaging and Communications in Medicine (DICOM) format.
It is worth noting that even MR images from the same dataset have intra-dataset variations (such as the matrix size, slice thickness, and number of slices).
Furthermore, inter-rater variability for the CG and PZ annotations exists, as different physicians delineated them.
For clinical feasibility~\cite{hoeks2011}, we analyzed only axial T2w MR slices---the most commonly used sequence for prostate zonal segmentation---among the available sequences.
In our multi-centric study, we conducted the following seven experiments resulting from all  possible training/testing conditions:
\begin{figure}[!t]
	\centering
	\subfloat[]{\includegraphics[height=3.95cm]{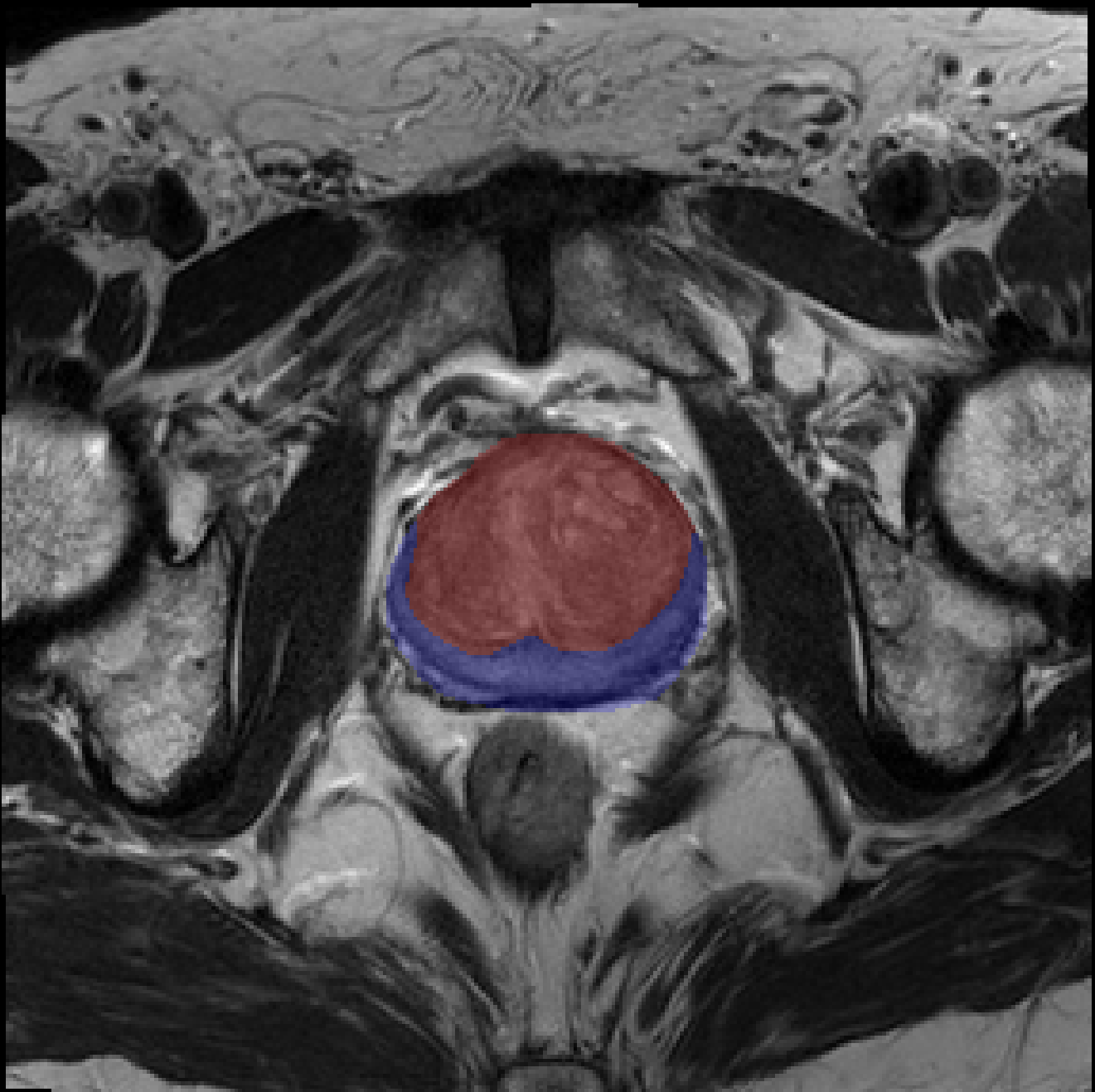}\label{fig:InputImagesA}}\quad
	\subfloat[]{\includegraphics[height=3.95cm]{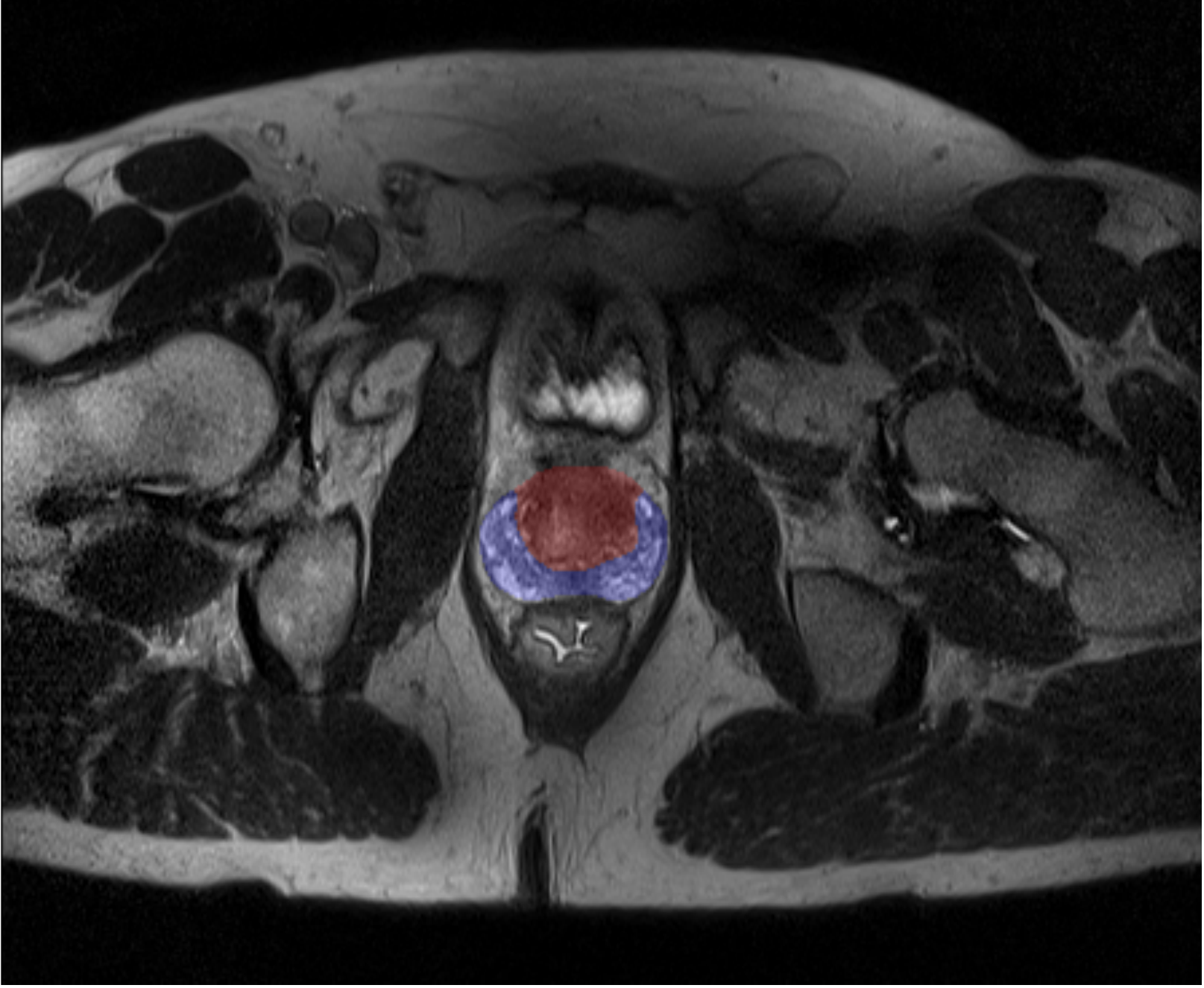}\label{fig:InputImagesB}}\quad
    \subfloat[]{\includegraphics[height=3.95cm]{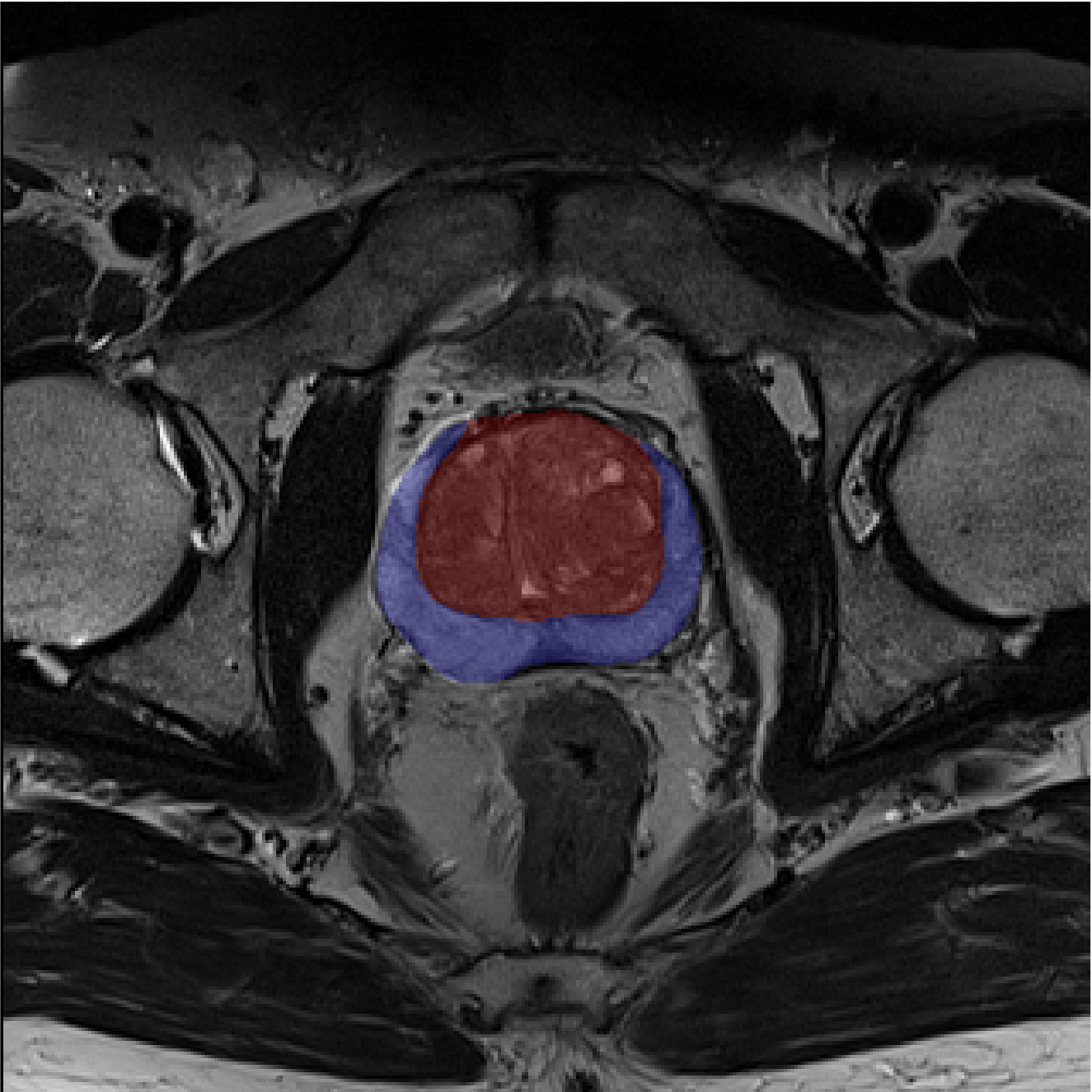}\label{fig:InputImagesC}}\\
	\caption{Examples of input prostate T2w MR axial slices in their original image ratio: (a) dataset $\#1$; (b) dataset $\#2$; (c) dataset $\#3$.
    The CG and PZ are highlighted with red and blue transparent regions, respectively. Alpha blending with $\alpha=0.2$.}
	\label{fig:InputImages}	
\end{figure}

\begin{itemize}
\item Individual dataset $\#1$, $\#2$, $\#3$: training and testing on dataset $\#1$ ($\#2$, $\#3$, respectively) alone in $4$-fold cross-validation, and testing also on whole datasets $\#2$ and $\#3$ ($\#1$ and $\#3$, $\#1$ and $\#2$, respectively) separately for each round;
\item Mixed dataset $\#1/\#2$, $\#2/\#3$, $\#1/\#3$: training and testing on both datasets $\#1$ and $\#2$ ($\#2$ and $\#3$, $\#1$ and $\#3$, respectively) in $4$-fold cross-validation, and testing also on whole dataset $\#3$ ($\#1$, $\#2$, respectively) separately for each round;
\item Mixed dataset $\#1/\#2/\#3$: training and testing on whole datasets $\#1$, $\#2$, and $\#3$ in $4$-fold cross-validation.
\end{itemize}

For clinical applications, such a multi-centric research is valuable for analyzing CNNs' generalization ability among different MRI acquisition options, e.g., different devices and functioning  parameters. In our study, for instance, both intra-/cross-scanner evaluations can be carried out, because dataset $\#1$'s scanner is different from those of datasets $\#2$ and $\#3$.
Fig.~\ref{fig:InputImages} shows an example image for each analyzed dataset; in the context of generalization among different datasets, Yan \textit{et al.}~\cite{yan2018} evaluated the average vessel segmentation performance on three retinal fundus image datasets under the three-dataset training condition, while pair-wisely assessing the cross-dataset performance on two datasets under the other one-dataset training condition. 
Yang \textit{et al.}~\cite{yang2018} proposed an alternative approach using adversarial appearance rendering to relieve the burden of re-training for Ultrasound imaging datasets. Differently, we thoroughly evaluate all possible training/testing conditions (for a total of $21$ configurations) on each dataset to confirm the intra- and cross-dataset generalization ability by incrementally injecting samples from the other datasets at hand.

With regard to the $4$-fold cross-validation, we partitioned the datasets $\#1$, $\#2$, and $\#3$ into $4$ folds by using the following patient indices: $\{ [1, \ldots, 5]$, $[6, \ldots, 10]$, $[11, \ldots, 15]$, \sloppy $[16, \ldots, 21] \}$, $\{ [1, \ldots, 5]$, $[6, \ldots, 10]$, $[11, \ldots, 15]$, $[16, \ldots, 19] \}$, and $\{ [1, \ldots, 10]$, $[11, \ldots, 20]$, $[21, \ldots, 30]$, $[31, \ldots, 40] \}$, respectively.
Finally, the results from the different cross-validation rounds were averaged to obtain a final descriptive value. These patient indices represent a permutation of the randomly arranged original patient ordering to portray a randomized partition scheme.
This allowed us to guarantee a fixed partitioning among the different training/testing conditions with a general notation valid for all datasets, regardless of the number of patients in each dataset.

Cross-validation strategies aim at estimating the generalization ability of a given model; the hold-out method fixedly partition the dataset into the training/test sets to train the model on the first partition alone and test it only on the unseen test set data.
Unlike the leave-one-out cross-validation with high variance and low bias, the $k$-fold cross-validation is a natural way to improve the hold-out method: the dataset is divided into $k$ mutually exclusive folds of approximately equal size \cite{diri2008}.
The statistical validity increases with less variance and less dependency on the initial dataset partition, averaging the results for all the $k$ cross-validation rounds.
Consequently, the $k$-fold cross-validation is the most common choice for reliable generalization results, minimizing the bias associated with the random sampling of the training/test sets \cite{diri2008}.
However, this statistical practice is computationally expensive due to the $k$ times-repeated training from scratch \cite{gandhi2010}.
Moreover, the results could underestimate the actual performance allowing for conservative analyses \cite{kohavi1995}; thus, we chose $4$-fold cross-validation for reliable and fair training/testing phases, according to the number of patients in each dataset, calculating the evaluation metrics on a statistically significant test set (i.e., $25\%$ of each prostate MRI dataset).

\subsection{Prostate Zonal Segmentation on Multi-institutional MRI Datasets}
\label{sec:Method}

This work adopts a selective delineation approach to focus on internal prostatic anatomy: the CG and PZ, denoted by $\mathcal{R}_{CG}$ and $\mathcal{R}_{PZ}$, respectively.
Let the entire image and the WG region be $\mathcal{I}_\Omega$ and $\mathcal{R}_{WG}$, respectively, the following relationships can be defined:
\begin{equation}
	\label{eq:globalConstraints}
	\mathcal{I}_\Omega = \mathcal{R}_{WG} \cup \mathcal{R}_{bg} \mbox{ and } \mathcal{R}_{WG} \cap \mathcal{R}_{bg} = \varnothing,
\end{equation}
where $\mathcal{R}_{bg}$ represents background pixels.
Relying on \cite{villeirs2007,qiu2014}, $\mathcal{R}_{PZ}$ was obtained by subtracting $\mathcal{R}_{CG}$ from $\mathcal{R}_{WG}$ meeting the constraints:
\begin{equation}
	\label{eq:segConstraints}
	\mathcal{R}_{WG} = \mathcal{R}_{CG} \cup \mathcal{R}_{PZ} \mbox{ and } \mathcal{R}_{CG} \cap \mathcal{R}_{PZ} = \varnothing.
\end{equation}

\begin{figure}[!t]
	\centering
	\includegraphics[width=\textwidth]{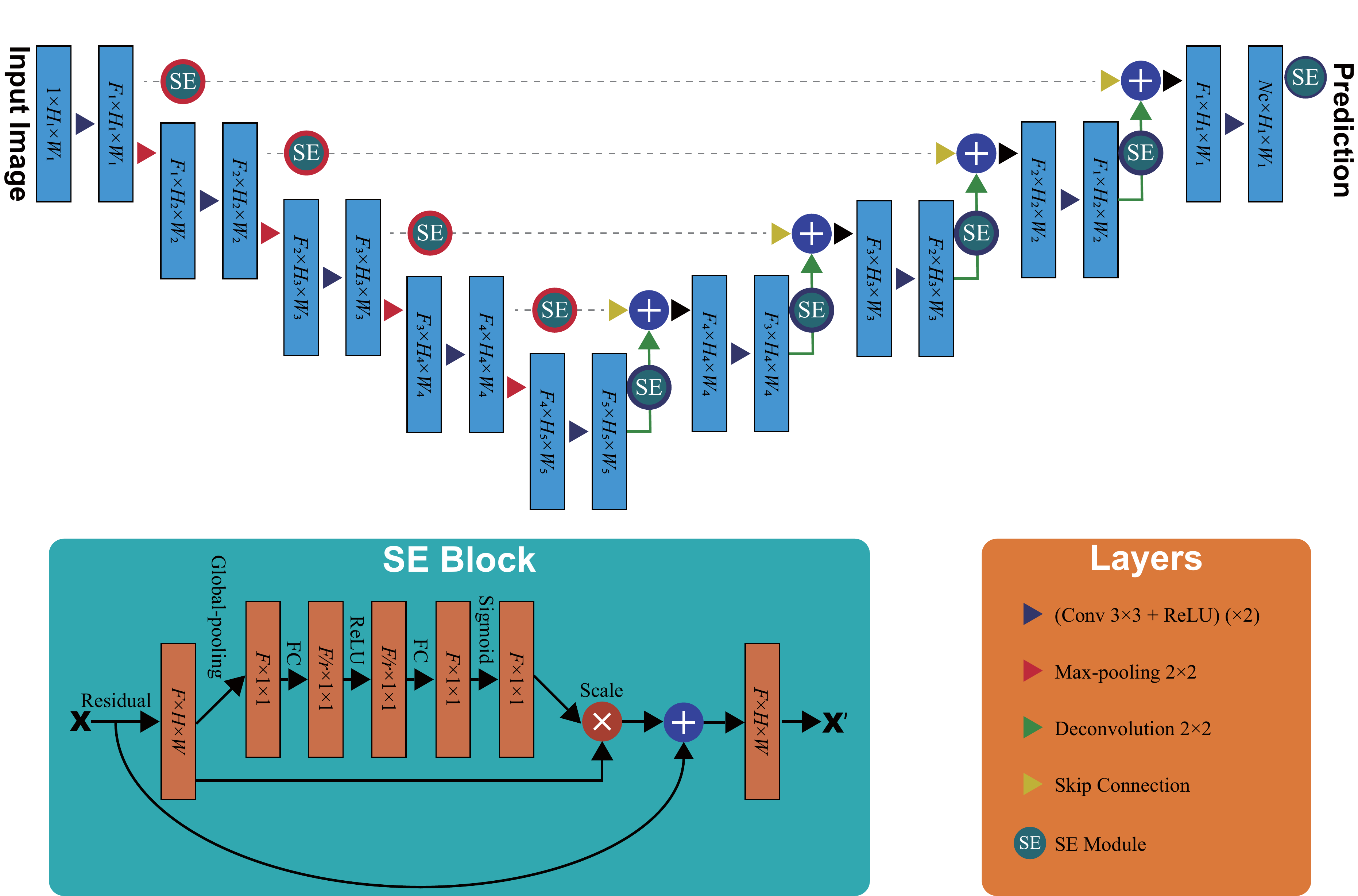}
	\caption{Scheme of the proposed USE-Net architecture: Enc USE-Net has only $4$ (red-contoured) SE blocks after every encoder, whilst Enc-Dec USE-Net has $9$ SE blocks integrated after every encoder/decoder (represented with red/blue contours, respectively).}
	\label{fig:usenet}
\end{figure}

\subsubsection{USE-Net: Incorporating SE Blocks into U-Net}
\label{sec:USEnet}

We propose to introduce SE blocks~\cite{hu2017} following every Encoder (Enc USE-Net) or Encoder-Decoder (Enc-Dec USE-Net) of U-Net~\cite{ronneberger2015}, as shown in Fig. \ref{fig:usenet}.
As pointed out before, U-Net allows for a multi-resolution decomposition/composition technique~\cite{suzuki2006}, by combining encoders/decoders with skip connections between them ~\cite{yao2018};
in our implementation, encoders and decoders consist of four pooling operators that capture the context and up-sampling operators that conduct precise localization, respectively.

We introduce SE blocks to enhance image segmentation, expecting an increased representational power from modeling the channel-wise dependencies of convolutional features~\cite{hu2017}.
These blocks were originally envisioned for image classification using adaptive feature recalibration to boost informative features and suppress the weak ones at minimal computational burden.

Enc USE-Net and Enc-Dec USE-Net are investigated to evaluate the effect of strengthened feature recalibration.
Since the template of the SE blocks is generic, they can be exploited at any depth of any architecture.
Considering that SE blocks should be placed after output feature maps for feature recalibration, we have three possible places to integrate them for U-Net, namely: (\textit{i}) after encoders; (\textit{ii}) after decoders; (\textit{iii}) after a classifier.
SE blocks are more powerful in the encoding path than in the decoding path and more powerful in the decoding path than after a classifier, as they affect lower-level features in the U-Net architecture and thus increase the overall performance significantly; consequently, instead of placing only a single SE block after the first encoder/decoder, we place SE blocks after each encoder/decoder for both coarse-grained context detection in the earlier layers and fine-grained localization in the deeper layers for the best segmentation performance.

The SE blocks can be formally described as follows:

\paragraph{Squeeze}
Let $\UU = [\uu_1, \uu_2, \dots, \uu_{F}]$ be an input feature map, where $\uu_f \in \mathbb{R}^{H \times W}$ is a single channel with size $H \times W$.
Through spatial dimensions $H \times W$, a global average pooling layer generates channel-wise statistics $\zz \in \RR^{F}$, whose $f$-th element is given by:
\begin{equation}\label{squeeze}
z_f = \frac{1}{H \times W}\sum_{h=1}^{H} \sum_{w=1}^{W} [\uu_f]_{i,j}.
\end{equation}

\paragraph{Excitation}
To limit the model complexity and boost generalization, two fully-connected layers and the Rectified Linear Unit (ReLU)~\cite{nair2010} function $\delta$ transform $\zz$ with a sigmoid activation function $\sigma(\cdot)$:
\begin{equation}\label{excitation}
\mathbf{s} = \sigma(g(\mathbf{z}, \mathbf{W})) = \sigma(\WW_2\delta(\WW_1\mathbf{z})),
\end{equation}
where $\WW_1 \in \RR^{\frac{F}{r} \times F}$, $\WW_2 \in \RR^{F \times \frac{F}{r}}$, and $r$ is the reduction ratio controling the capacity and computational cost of the SE blocks.
Hu \textit{et al.}~\cite{hu2017} showed that the SE blocks can overfit to the channel inter-dependencies of the training set despite a lower number of weights with respect to the original architecture; they found the best compromise of $r = 8$, which guarantees the lowest overall error (in terms of top-$1$ and top-$5$ errors) with ResNet-50~\cite{he2016} for the ImageNet Large Scale Visual Recognition Challenge (ILSVRC) 2017 classification competition \cite{ILSVRC15}.
Therefore, we also selected $r = 8$ for the USE-Net.
In order to obtain an adaptive recalibration that ignores less important channels and emphasizes important ones (allowing for non-mutual exclusivity among multiple channels, differently from one-hot encoding), $\UU$ is rescaled into $\widetilde{\XX} = [\widetilde{\xx}_1, \widetilde{\xx}_2, \dots, \widetilde{\xx}_{F}]$ by applying Eq.~(\ref{eq:Fscale}):
\begin{equation}\label{eq:Fscale}
\widetilde{\mathbf{x}}_f = \mathbf{F}_\text{scale}(\mathbf{u}_f, s_f) = s_f \cdot \mathbf{u}_f, \mbox{ for } f = 1,2, \ldots, F,
\end{equation}
where $\FF_\text{scale}(\uu_f, s_f)$ represents the channel-wise multiplication between the feature map $\uu_f \in \RR^{H \times W}$ and the scalar $s_f \in [0,1]$.

\subsubsection{Pre-processing}
\label{sec:PreProc}
To fit the image resolution of dataset $\#1$, we either center-cropped or zero-padded the images of datasets $\#2$ and $\#3$ to resize them to $288 \times 288$ pixels.
Afterwards, all images in the three datasets were masked using the corresponding prostate binary masks to omit the background and only focus on extracting the CG and PZ from the WG.
This operation can be performed either by an automated method~\cite{rundo2017Inf} or previously provided manual WG segmentation~\cite{lemaitre2015}.
As a simple form of data augmentation, we randomly cropped the input images from $288 \times 288$ to $256 \times 256$ pixels and horizontally flipped them.

\subsubsection{Post-processing}
\label{sec:PostProc}

Two efficient morphological operations were applied on the obtained $\mathcal{R}_{CG}$ binary masks to smooth boundaries and deal with disconnected regions:
\begin{itemize}
	\item a hole filling algorithm on the segmented $\mathcal{R}_{CG}$ to remove possible holes in a predicted map;
    \item  a small area removal operation dealing with connected components smaller than $\lfloor |\mathcal{R}_{WG}|/8 \rfloor$ pixels, where $|\mathcal{R}_{WG}|$ denotes the number of pixels contained in WG segmentation.
This adaptive criterion takes into account the different sizes of $\mathcal{R}_{WG}$ (ranging from the apical to the basal prostate slices).
\end{itemize}

\subsubsection{Comparison against the State-of-the-Art Methods}
\label{sec:CompSotA}

We compare USE-Net against three supervised CNN-based architectures (i.e., U-Net, pix2pix, and Mixed-Scale Dense Network) and the unsupervised continuous max-flow model~\cite{qiu2014}.
All the investigated CNN-based architectures were trained using the $\mathcal{L}_{DSC}$ loss function (i.e., a continuous version of the \textit{DSC})~\cite{milletari2016} through the  $N$ pixels to classify:
\begin{equation}
	\label{eq:DSCloss}
	\mathcal{L}_{DSC} = - \frac{2\sum_{i=1}^{N} s_i \cdot r_i}{\sum_{i=1}^{N} s_i + \sum_{i=1}^{N} r_i},
\end{equation}
where $s_i$ and $r_i$ refer to the continuous values in $[0, 1]$ of the prediction map and the Boolean ground truth annotated by experienced radiologists at the $i$-th pixel, respectively.
The $\mathcal{L}_{DSC}$ loss function was designed by Milletari \textit{et al.}~\cite{milletari2016} to deal with the imbalance of the foreground labels in medical image segmentation tasks.

\paragraph{USE-Net and U-Net}
Using four scaling operations, U-Net and USE-Net were implemented on Keras with TensorFlow backend.
We used the Stochastic Gradient Descent (SGD) method~\cite{bottou2010} with a learning rate of $0.01$, momentum of $0.9$, weight decay of $5 \times 10^{-4}$, and batch size of $4$.
Training was executed for $50$ epochs, multiplying the learning rate by $0.2$ at the $20$-th and $40$-th epochs.

\paragraph{pix2pix}
  This image-to-image translation method with conditional adversarial networks was used to translate the original image into the segmented one~\cite{isola2016}.
The generator and discriminator (both U-Nets in our implementation) include eight and five scaling operations, respectively.
We developed pix2pix on PyTorch. Adam~\cite{kingma2014} was used as an optimizer with a learning rate of $0.01$ for the generator---which was multiplied by $0.1$ every $20$ epochs---and $2 \times 10^{-4}$ for the discriminator. 
Training was executed for $50$ epochs with a batch size of $12$.

\paragraph{MS-D Net}
This dilated convolution-based method, characterized by densely connected feature maps, is designed to capture features at various image scales \cite{pelt2017}.
It was implemented on PyTorch with a depth of $100$ and width of $1$.
We used Adam~\cite{kingma2014} with a learning rate of $1 \times 10^{-3}$ and trained it for $100$ epochs with a batch size of $12$.

\paragraph{Continuous Max-flow Model}
This model~\cite{qiu2014} exploits duality-based convex relaxed optimization~\cite{yuan2010} to achieve better numerical stability (i.e., convergence) than classic graph cut-based methods~\cite{freedman2005}.
This semi-automatic approach simultaneously segments both $\mathcal{R}_{WG}$ and $\mathcal{R}_{CG}$ under the constraints given in Eq.~(\ref{eq:segConstraints}), relying on user intervention.
The initialization procedure consists in two closed surfaces defined by a thin-plate spline interpolating $10$-$12$ control points interactively selected by the user (considering both the axial and sagittal views).
These 3D partitions estimate the intensity probability density functions associated with three sub-regions of background, CG, and PZ.
This allows for defining the region appearance models for global optimization-based multi-region segmentation~\cite{yuan2010}.

Since the supervised CNN-based architectures rely on the gold standard $\mathcal{R}_{WG}$ for zonal segmentation, we apply the continuous max-flow method on CG for single-region segmentation for a fair comparison.
Moreover, in our tests, a very accurate slice-by-slice $\mathcal{R}_{CG}$ initialization is provided by eroding the gold standard CG with a circular structuring element (radius $= 6$ pixels).

The continuous max-flow model~\cite{qiu2014} was implemented in MatLab\textsuperscript{\textregistered} R2017a $64$-bit (The Mathworks, Natick, MA, USA).

\vspace{0.03in}

\subsection{Evaluation Metrics}
\label{sec:Eval}
We evaluate the segmentation methods by comparing the
segmented MR images ($\mathcal{S}$) to the corresponding gold standard manual segmentation ($\mathcal{G}$) using spatial overlap- and distance-based metrics \cite{taha2015,fenster2005,zhang2001}.
Those metrics are calculated using a slice-wise comparison and then averaged per patient;
thus, each single result regarding a patient represents an aggregate value.

\paragraph{Overlap-based metrics}
These metrics quantify the spatially-overlapping segmented Region of Interest (ROI).
Let true positives be $TP = \mathcal{S} \cap \mathcal{G}$, false negatives be $FN = \mathcal{G} - \mathcal{S}$, false positives be $FP = \mathcal{S} - \mathcal{G}$, and true negatives be $TN = \mathcal{I}_\Omega - \mathcal{G} - \mathcal{S}$.
In what follows, we denote the cardinality of the pixels belonging to a region $\mathcal{A}$ as $|\mathcal{A}|$.

\begin{itemize}
	\setlength\itemsep{1em}
	\item \textit{Dice similarity coefficient}~\cite{zou2004} is the most used measure in medical image segmentation to compare the overlap of two regions:
	\begin{equation}
	DSC = \frac{2 \cdot \abs{TP}}{\abs{\mathcal{S}} + \abs{\mathcal{G}}} \cdot 100.
	\end{equation}
	\item \textit{Sensitivity} measures the correct detection ratio of true positives:
	\begin{equation}
	SEN = \frac{\abs{TP}} {\abs{TP} + \abs{FN}} \cdot 100.
	\end{equation}
	\item \textit{Specificity} measures the correct detection ratio of true negatives:
	\begin{equation}
	\label{eq:TNR}
	TNR = \frac{\abs{TN}}{\abs{TN} + \abs{FP}} \cdot 100.
	\end{equation}
	However, this formulation is ineffective when data are unbalanced (i.e., the ROI is much smaller than the whole image). Consequently, we use the following definition:
	\begin{equation}
	\label{eq:SPC}
	SPC = \left(1 - \frac{\abs{FP}}{\abs{\mathcal{S}}} \right)  \cdot 100.
	\end{equation}	
\end{itemize}

\paragraph{Distance-based metrics}
As precise boundary tracing plays an important role in clinical practice, overlap-based metrics have limitations in evaluating segmented images.
In order to measure the distance between the two ROI boundaries, distance-based metrics can be considered.
Let the manual contour $G$ consist in a set of vertices $\{ \mathbf{g}_a: a = 1, 2, \dots, A \}$ and the automatically-generated contour $S$ consist in a set of vertices $\{ \mathbf{s}_b: b = 1, 2, \dots, B \}$.
We calculate the absolute distance between an arbitrary element $\mathbf{s}_b \in S$ and all the vertices in $G$ as follows:
\begin{equation}d(\mathbf{s}_b, G) = \min_{a \in \{ 1,2, \dots, A \}} \| \mathbf{s}_b - \mathbf{g}_a \|.
\end{equation}

\begin{itemize}
	\setlength\itemsep{1em}
	\item \textit{Average absolute distance} measures the average difference between the ROI boundaries of $\mathcal{S}$ and $\mathcal{G}$:
	\begin{equation}
	AvgD = \frac{1}{B}\sum \limits_{b=1}^B d(\mathbf{s}_b, G).
	\end{equation}
	\item \textit{Maximum absolute distance} represents the maximum difference between the ROI boundaries of $\mathcal{S}$ and $\mathcal{G}$:
	\begin{equation}
	MaxD = \max_{b \in \{ 1, 2, \dots, B \}}  d(\mathbf{s}_b, G).
	\end{equation}
\end{itemize}

\section{Experimental Results}
\label{sec:Results}
This section shows how the CNN-based architectures and the continuous max-flow model segmented the prostate zones, through the evaluation of their cross-dataset generalization ability.
Aiming at showing the performance boost achieved by integrating the SE blocks into U-Net, we performed a fair comparison against the state-of-the-art architectures under the same training/testing conditions.
In particular, due to the lack of annotated MR images for prostate zonal segmentation, we used three different datasets by composing a multi-institutional dataset.
This allowed us to show the SE blocks' cross-dataset adaptive feature recalibration effect, better capturing each dataset's peculiar characteristics.
Therefore, we exploited all possible training/testing conditions involving the three analyzed datasets (for a total of $21$ configurations) on each dataset to overcome the limitation from the small sample size, confirming the intra- and cross-dataset generalization ability of the CNN-based architectures.

\begin{figure}[!t]
\centering
	\subfloat[][]
    {\includegraphics[width=\textwidth]{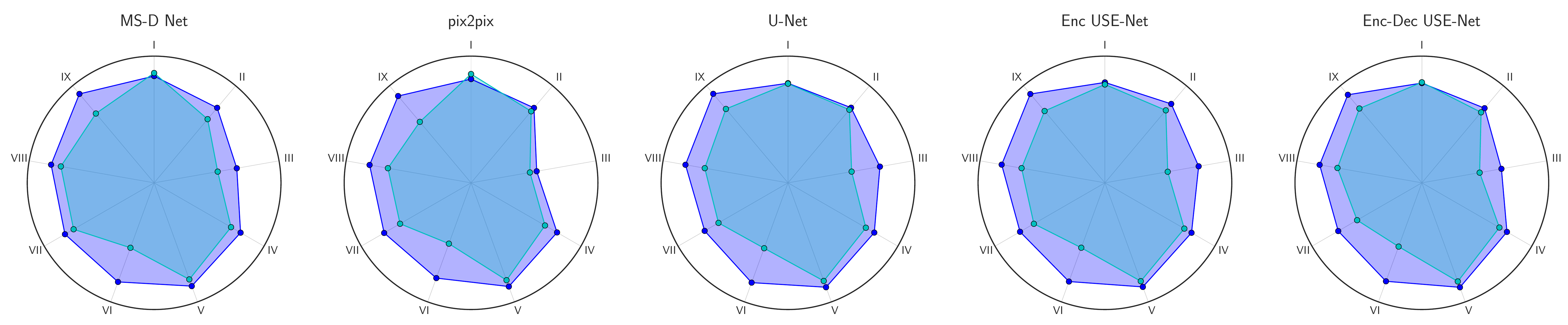}} \\
    \subfloat[][]
    {\includegraphics[width=\textwidth]{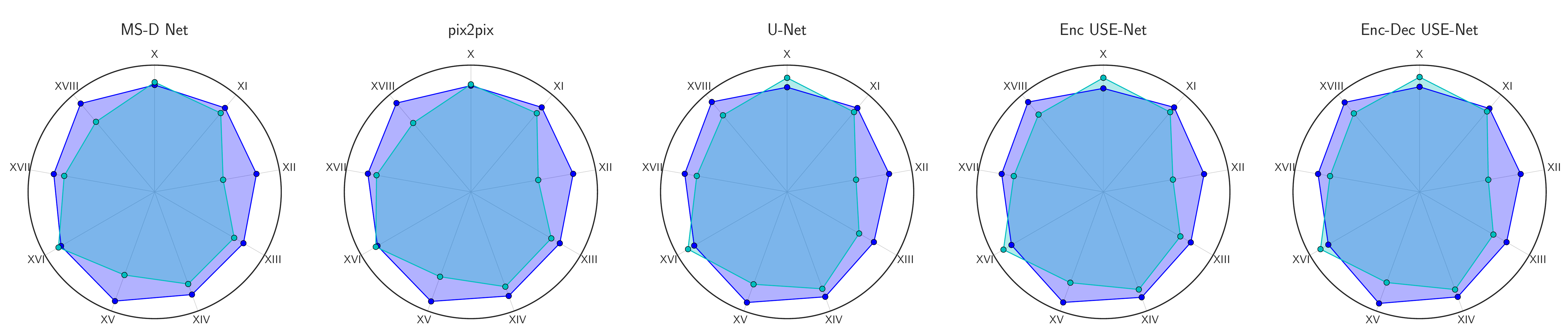}} \\
     \subfloat[][]
    {\includegraphics[width=\textwidth]{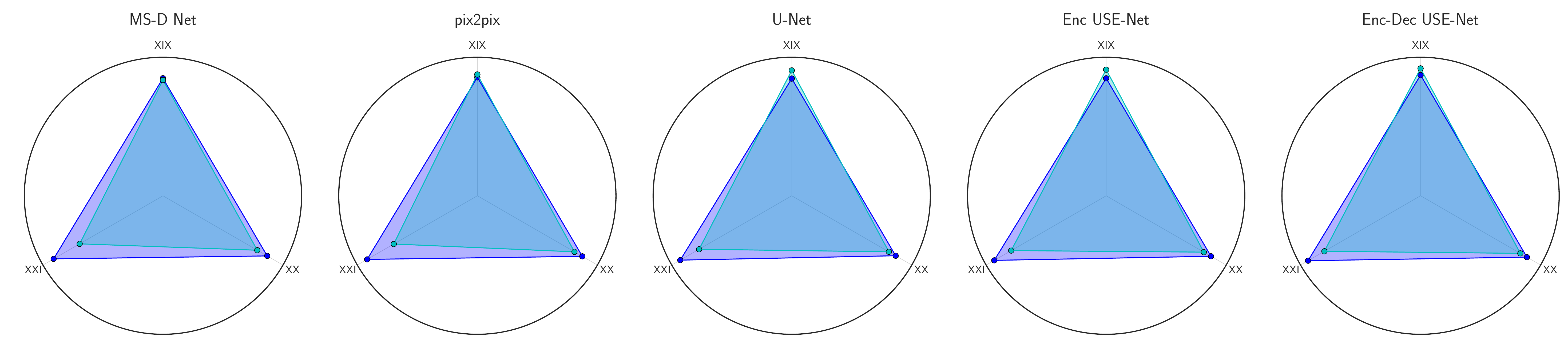}} \\
  \caption{Kiviat diagrams showing the \textit{DSC} values achieved by each method under different conditions.
  $\mathcal{R}_{CG}$ and $\mathcal{R}_{PZ}$ results are denoted by blue and cyan colors, respectively. Each variable represents a ``training-set $\rightarrow$ test-set'' condition as follows:\newline
(a) one-dataset training: I) $\#1\rightarrow\#1$; II) $\#1\rightarrow\#2$; III) $\#1\rightarrow\#3$; IV) $\#2\rightarrow\#1$; V) $\#2\rightarrow\#2$; VI) $\#2\rightarrow\#3$; VII) $\#3\rightarrow\#1$; VIII) $\#3\rightarrow\#2$; IX) $\#3\rightarrow\#3$.\newline
(b) two-dataset training: X) $\#1/\#2\rightarrow\#1$; XI) $\#1/\#2\rightarrow\#2$; XII) $\#1/\#2\rightarrow\#3$; XIII) $\#1/\#3\rightarrow\#1$; XIV) $\#1/\#3\rightarrow\#2$; XV) $\#1/\#3\rightarrow\#3$; XVI) $\#2/\#3\rightarrow\#1$; XVII) $\#2/\#3\rightarrow\#2$; XVIII) $\#2/\#3\rightarrow\#3$. \newline 
(c) three-dataset training: XIX) $\#1/\#2/\#3\rightarrow\#1$; XX) $\#1/\#2/\#3\rightarrow\#2$; XXI) $\#1/\#2/\#3\rightarrow\#3$.}
\label{fig:Kiviat}
\end{figure}

Table \ref{table:results} shows the $4$-fold cross-validation results, as assessed by the \textit{DSC} metrics, obtained under different training/testing conditions (the values of the other metrics are given in Supplementary Material, Tables S1-S4).
For visual and comprehensive comparison, the Kiviat diagrams (also known as radar or cobweb charts)~\cite{kolence1973,diri2008} for each CNN-based architecture are also displayed in Fig. \ref{fig:Kiviat}.
Here, we can observe the impact of leaving dataset $\#3$ out of the training set and, at the same time, using it as test set: the corresponding spokes III, VI, and XII generally show lower performance, probably due to the peculiar image characteristics of dataset $\#3$ (comprising the highest number of patients) that are not learned during the training phase on datasets $\#1/\#2$.
In general, Enc USE-Net performs similarly to U-Net, which stably yields satisfactory results.
More interestingly, Enc USE-Net obtains considerably better results when trained/tested on multiple datasets.
Enc-Dec USE-Net (characterized by a higher number of SE blocks with respect to Enc USE-Net) consistently and remarkably outperforms the other methods on both CG and PZ segmentation when trained on all the investigated datasets, also performing well when trained and tested on the same datasets.

\begin{table*}[!t]
\tiny
\centering
  \caption{Prostate zonal segmentation results of the CNN-based architectures and the unsupervised continuous max-flow model (proposed by Qiu~\textit{et al.}~\cite{qiu2014}) in $4$-fold cross-validation assessed by \emph{DSC} (presented as the mean value $\pm$ standard deviation). The supervised experimental results are calculated under the different seven conditions described in Section \ref{sec:Datasets}. Numbers in bold indicate the best \emph{DSC} values (the higher the better) for each prostate region (i.e., $\mathcal{R}_{CG}$ and $\mathcal{R}_{PZ}$) among all architectures.\vspace{0.2cm}}
  
\label{table:results}
\begin{tabular}{p{3.3em}|l|ll|ll|ll}
\Hline
\multirow{2}{*}{\hspace{30pt}}   & \multicolumn{1}{c|}{\multirow{2}{*}{\textbf{Method}}} & \multicolumn{2}{c|}{\textbf{Testing on dataset $\#1$}} & \multicolumn{2}{c|}{\textbf{Testing on dataset $\#2$}} & \multicolumn{2}{c}{\textbf{Testing on dataset $\#3$}}\\
                            & \multicolumn{1}{c|}{}                                      & \multicolumn{1}{c}{\textit{CG}}            & \multicolumn{1}{c|}{\textit{PZ}}            & \multicolumn{1}{c}{\textit{CG}}          & \multicolumn{1}{c|}{\textit{PZ}} & \multicolumn{1}{c}{\textit{CG}}          & \multicolumn{1}{c}{\textit{PZ}}       \\ \hline

\parbox[t]{2mm}{\multirow{5}{*}{\rotatebox[origin=c]{270}{\textbf{\shortstack{\\Training on\\dataset\\ $\#1$\vspace{0.4mm}}}}}} & MS-D Net
                                        &  $\textbf{84.3}\pm1.6$                &       $\textbf{86.7}\pm1.6$               &       $77.3\pm4.1$           &     $65.6\pm11.2$ &       $66.2\pm3.7$           &     $\textbf{50.9}\pm1.2$             \\

                            & pix2pix                                        &  $81.9\pm2.2$                &       $85.9\pm5.0$               &       $77.2\pm2.9$           &     $73.7\pm4.3$ &       $52.5\pm3.2$           &     $47.1\pm1.3$             \\
                            & U-Net                                        &  $78.6\pm4.1$                &       $78.3\pm7.6$               &       $77.4\pm5.4$           &     $\textbf{75.3}\pm1.4$ &       $73.6\pm6.2$           &     $50.9\pm1.5$             \\

& Enc USE-Net                                        &  $79.3\pm3.5$                &       $77.7\pm2.7$               &       $\textbf{81.3}\pm1.5$           &     $74.7\pm1.8$ &       $\textbf{75.0}\pm4.2$           &     $50.3\pm1.2$             \\
                            & Enc-Dec USE-Net                                        &  $78.8\pm2.9$                &       $79.4\pm7.9$               &       $76.9\pm5.5$           &     $72.7\pm1.7$ &       $63.7\pm14.6$           &     $46.3\pm1.8$             \\
\hline
\parbox[t]{2mm}{\multirow{5}{*}{\rotatebox[origin=c]{270}{\textbf{\shortstack{\\Training on\\dataset\\ $\#2$\vspace{0.4mm}}}}}} & MS-D Net
                                        &  $78.7\pm1.1$                &       $70.0\pm4.4$               &       $86.8\pm3.7$           &     $81.1\pm0.5$ &       $83.2\pm1.0$           &     $54.6\pm0.8$             \\

                            & pix2pix                                        &  $78.3\pm0.9$                &       $67.3\pm3.2$               &       $87.1\pm2.9$           &     $81.8\pm1.0$ &       $80.0\pm2.5$           &     $51.1\pm1.5$             \\
                            & U-Net                                        &  $78.6\pm1.0$                &       $70.9\pm3.2$               &       $87.7\pm2.0$           &     $82.4\pm2.4$ &       $\textbf{83.8}\pm1.8$           &     $\textbf{54.9}\pm1.8$             \\

& Enc USE-Net                                        &  $\textbf{78.8}\pm1.4$                &       $\textbf{72.3}\pm5.6$               &       $87.4\pm2.5$           &     $82.6\pm2.1$ &       $82.9\pm2.5$           &     $54.5\pm2.0$             \\

                            & Enc-Dec USE-Net                                        &  $77.5\pm2.1$                &       $70.6\pm5.5$               &       $\textbf{87.8}\pm2.7$           &     $\textbf{82.8}\pm1.9$ &       $82.7\pm1.5$           &     $53.6\pm1.0$             \\
\hline
\parbox[t]{2mm}{\multirow{5}{*}{\rotatebox[origin=c]{270}{\textbf{\shortstack{\\Training on\\dataset\\$\#3$\vspace{0.4mm}}}}}} & MS-D Net
                                        &  $\textbf{81.2}\pm1.3$                &       $\textbf{73.3}\pm3.7$               &       $82.5\pm1.9$           &     $\textbf{74.7}\pm2.0$ &       $91.6\pm1.1$           &     $71.4\pm5.6$             \\

                            & pix2pix                                        &  $79.1\pm5.6$                &       $64.6\pm22.1$               &       $81.2\pm4.2$           &     $66.6\pm19.1$ &       $89.4\pm4.8$           &     $62.8\pm10.0$             \\
                            & U-Net                                        &  $75.9\pm3.4$                &       $63.3\pm5.0$               &       $82.1\pm2.9$           &     $66.6\pm8.4$ &       $\textbf{91.7}\pm2.4$           &     $76.1\pm4.1$             \\

& Enc USE-Net                                        &  $77.3\pm3.6$                &       $64.7\pm6.4$               &       $\textbf{82.7}\pm4.3$           &     $66.7\pm15.9$ &       $91.5\pm3.2$           &     $74.0\pm7.8$             \\

                            & Enc-Dec USE-Net                                        &  $76.1\pm4.2$                &       $58.9\pm13.7$               &       $81.8\pm4.8$           &     $67.6\pm13.2$ &       $90.7\pm3.1$           &     $\textbf{76.6}\pm7.8$             \\
\hline \hline
\parbox[t]{2mm}{\multirow{5}{*}{\rotatebox[origin=c]{270}{\textbf{\shortstack{\\Training on\\datasets\\$\#1/\#2$\vspace{0.4mm}}}}}} & MS-D Net
                                        &  $84.4\pm3.1$                &       $86.5\pm2.7$               &       $86.4\pm2.8$           &     $81.2\pm1.3$ &       $81.7\pm2.3$           &     $54.9\pm2.5$             \\

                            & pix2pix                                        &  $\textbf{83.8}\pm2.6$                &       $84.8\pm3.1$               &       $\textbf{87.1}\pm2.7$           &     $81.0\pm0.4$ &       $\textbf{82.1}\pm2.5$           &     $54.0\pm1.8$             \\
                            & U-Net                                        &  $82.6\pm3.3$                &       $90.0\pm2.7$               &       $86.4\pm2.0$           &     $82.2\pm2.7$ &       $81.8\pm2.1$           &     $55.3\pm2.5$             \\

& Enc USE-Net                                        &  $81.7\pm5.4$                &       $90.0\pm2.1$               &       $87.0\pm2.1$           &     $82.2\pm1.8$ &       $80.8\pm2.7$           &     $\textbf{55.8}\pm1.7$             \\

                            & Enc-Dec USE-Net                                        &  $82.9\pm3.4$                &       $\textbf{90.6}\pm1.8$               &       $85.9\pm2.1$           &     $\textbf{82.9}\pm1.4$ &       $81.1\pm2.7$           &     $55.1\pm2.1$             \\
\hline
\parbox[t]{2mm}{\multirow{5}{*}{\rotatebox[origin=c]{270}{\textbf{\shortstack{\\Training on\\datasets\\$\#1/\#3$\vspace{0.4mm}}}}}} & MS-D Net
                                        &  $\textbf{85.4}\pm1.8$                &       $87.7\pm2.5$               &       $80.9\pm2.7$           &     $72.6\pm3.7$ &       $91.0\pm2.9$           &     $72.2\pm1.9$             \\

                            & pix2pix                                        &  $85.2\pm1.6$                &       $86.8\pm2.4$               &       $\textbf{82.7}\pm1.9$           &     $\textbf{75.7}\pm3.6$ &       $91.5\pm1.9$           &     $71.0\pm3.6$             \\
                            & U-Net                                        &  $84.8\pm0.4$                &       $90.4\pm2.8$               &       $82.1\pm2.9$           &     $72.5\pm4.5$ &       $\textbf{92.6}\pm1.5$           &     $78.9\pm4.0$             \\

& Enc USE-Net                                        &  $83.8\pm1.4$                &       $\textbf{91.1}\pm1.4$               &       $81.6\pm3.7$           &     $71.9\pm8.1$ &       $92.5\pm1.9$           &     $79.6\pm2.1$             \\

                            & Enc-Dec USE-Net                                        &  $83.3\pm3.2$                &       $90.4\pm2.1$               &       $81.5\pm4.6$           &     $71.8\pm6.7$ &       $92.2\pm2.4$           &     $\textbf{80.8}\pm1.8$             \\
\hline
\parbox[t]{2mm}{\multirow{5}{*}{\rotatebox[origin=c]{270}{\textbf{\shortstack{\\Training on\\datasets\\$\#2/\#3$\vspace{0.4mm}}}}}} & MS-D Net
                                        &  $81.0\pm1.3$                &       $72.5\pm5.6$               &       $86.2\pm2.5$           &     $77.4\pm4.7$ &       $91.7\pm0.9$           &     $69.8\pm3.6$             \\

                            & pix2pix                                        &  $\textbf{81.1}\pm1.1$                &       $\textbf{73.4}\pm3.3$               &       $87.4\pm2.2$           &     $79.6\pm5.7$ &       $92.0\pm1.3$           &     $71.3\pm3.4$             \\
                            & U-Net                                        &  $79.2\pm2.0$                &       $65.7\pm6.3$               &       $88.1\pm2.9$           &     $81.4\pm2.6$ &       $92.9\pm1.1$           &     $\textbf{77.6}\pm3.0$             \\

& Enc USE-Net                                        &  $79.8\pm1.8$                &       $70.3\pm7.6$               &       $\textbf{88.5}\pm2.4$           &     $\textbf{82.0}\pm3.2$ &       $92.8\pm1.0$           &     $76.3\pm2.7$             \\

                            & Enc-Dec USE-Net                                        &  $79.4\pm2.5$                &       $67.4\pm8.9$               &       $88.2\pm2.9$           &     $82.0\pm4.1$ &       $\textbf{93.7}\pm0.6$           &     $76.1\pm3.4$             \\
\hline \hline
\parbox[t]{2mm}{\multirow{5}{*}{\rotatebox[origin=c]{270}{\textbf{\shortstack{\\Training on\\datasets\\$\#1/\#2/\#3$\vspace{0.4mm}}}}}} & MS-D Net
                                        &  $84.8\pm4.5$                &       $83.6\pm6.9$               &       $86.8\pm2.6$           &     $78.6\pm5.1$ &       $91.1\pm1.0$           &     $69.4\pm4.5$             \\

                            & pix2pix                                        &  $85.5\pm2.6$                &       $87.6\pm3.5$               &       $87.5\pm2.0$           &     $80.9\pm5.3$ &       $91.8\pm1.3$           &     $69.7\pm4.8$             \\
                            & U-Net                                        &  $84.6\pm1.9$                &       $90.5\pm3.0$               &       $86.6\pm2.0$           &     $80.9\pm3.3$ &       $92.9\pm1.1$           &     $77.2\pm2.0$             \\

& Enc USE-Net                                        &  $84.8\pm2.3$                &       $91.1\pm2.5$               &       $87.4\pm1.8$           &     $81.4\pm4.4$ &       $93.2\pm0.7$           &     $79.1\pm3.5$             \\

                            & Enc-Dec USE-Net                                        &  $\textbf{87.1}\pm3.6$                &       $\textbf{91.9}\pm2.1$               &       $\textbf{88.6}\pm1.5$           &     $\textbf{83.1}\pm2.9$ &       $\textbf{93.7}\pm1.0$           &    $\textbf{80.1}\pm5.5$             \\
\hline \hline
\textbf{\,\,None} & Qiu \textit{et al.} \cite{qiu2014}
                                        &  $78.0\pm4.9$                &       $75.3\pm6.4$               &       $71.0\pm7.0$           &     $77.3\pm2.6$ &       $82.1\pm1.5$           &     $61.9\pm4.6$             \\
\Hline
\end{tabular}
\end{table*}


\begin{figure}[!t]
\captionsetup[subfigure]{labelformat=empty}
\centering
	\subfloat[]{\includegraphics[width=0.85\textwidth]{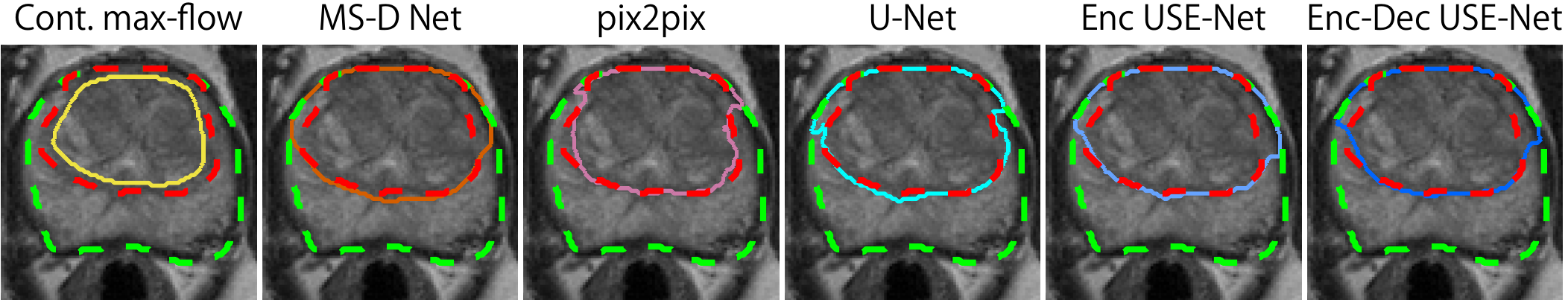}\label{sfig:segResD1a}}\\
    \addtocounter{subfigure}{-1}
    \captionsetup[subfigure]{labelformat=parens}
    \vspace{-1.1cm}
	\subfloat[]{\includegraphics[width=0.85\textwidth]{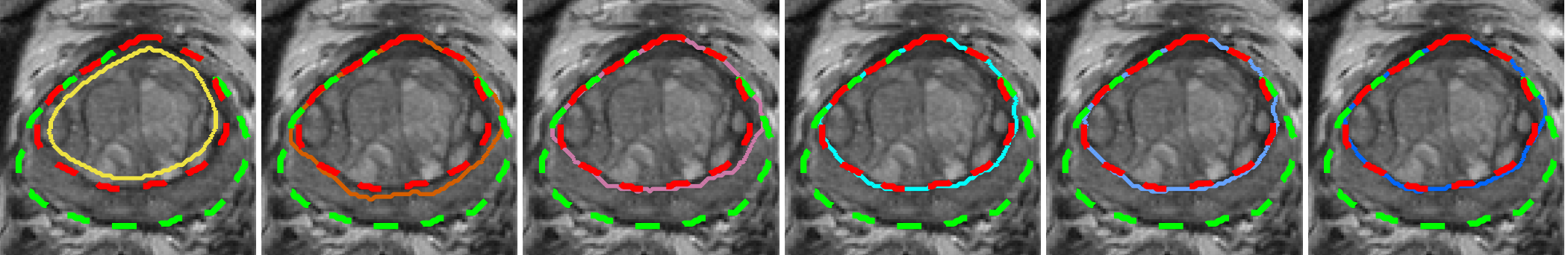}\label{sfig:segResD1b}}\\
    \captionsetup[subfigure]{labelformat=empty}
    \vspace{-0.4cm}
    	\subfloat[]{\includegraphics[width=0.85\textwidth]{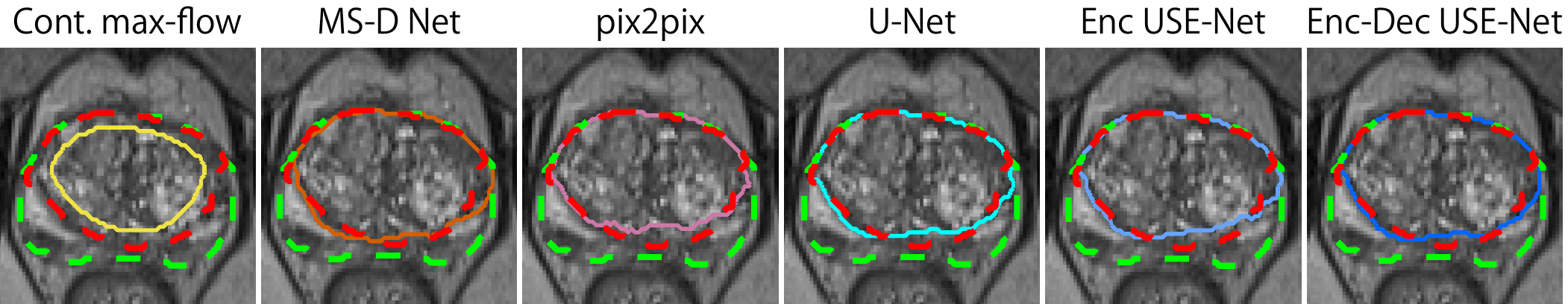}\label{sfig:segResD2a}}\\
        \addtocounter{subfigure}{-1}
        \captionsetup[subfigure]{labelformat=parens}
    \vspace{-1.1cm}
	\subfloat[]{\includegraphics[width=0.85\textwidth]{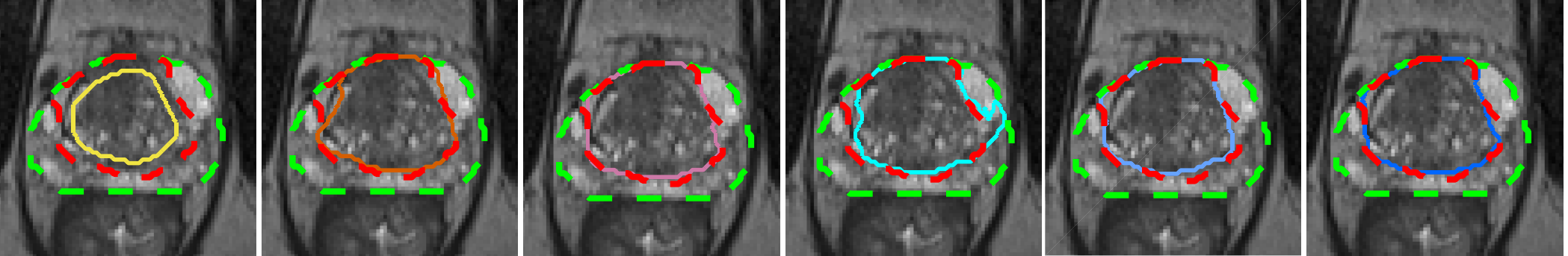}\label{sfig:segResD2b}}\\
    \captionsetup[subfigure]{labelformat=empty}
    \vspace{-0.4cm}
    	\subfloat[]{\includegraphics[width=0.85\textwidth]{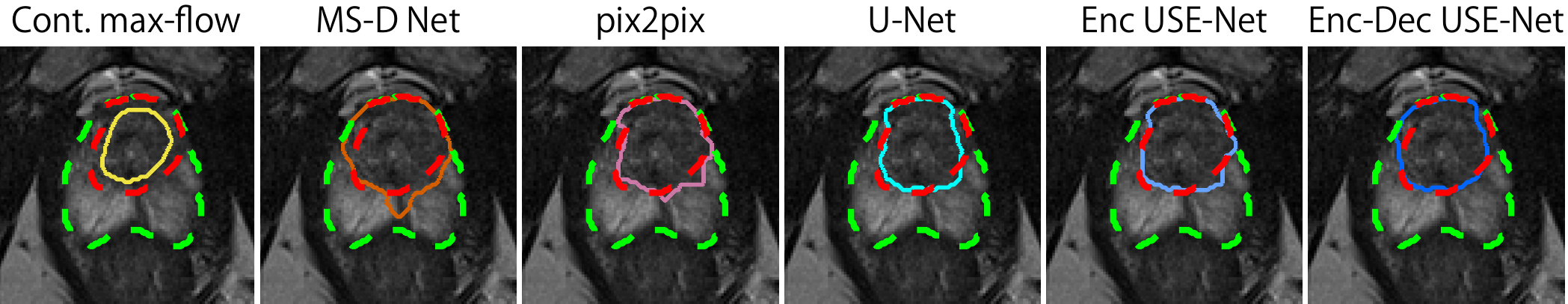}\label{sfig:segResD3a}}\\
        \addtocounter{subfigure}{-1}
        \captionsetup[subfigure]{labelformat=parens}
    \vspace{-1.1cm}
	\subfloat[]{\includegraphics[width=0.85\textwidth]{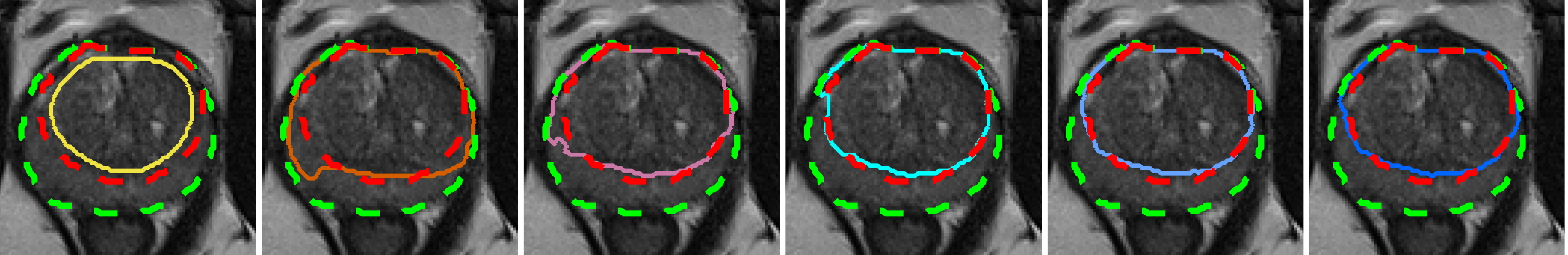}\label{sfig:segResD3b}}\\
	\caption{Segmentation results obtained by the six investigated methods (under the three-dataset training condition) on two different images for each dataset: (a) $\#1$; (b) $\#2$; (c) $\#3$. Automatic $\mathcal{R}_{CG}$ segmentations (solid lines) are compared against the corresponding gold standards (dashed red line). $\mathcal{R}_{PZ}$ segmentations can be obtained from $\mathcal{R}_{CG}$ and $\mathcal{R}_{WG}$ (dashed green line) according to the constraints in Eq.~(\ref{eq:segConstraints}).}
	\label{fig:segResD}
\end{figure}

\begin{figure}[t!]
\centering
	\subfloat[][Central Gland]
    {\includegraphics[width=\textwidth]{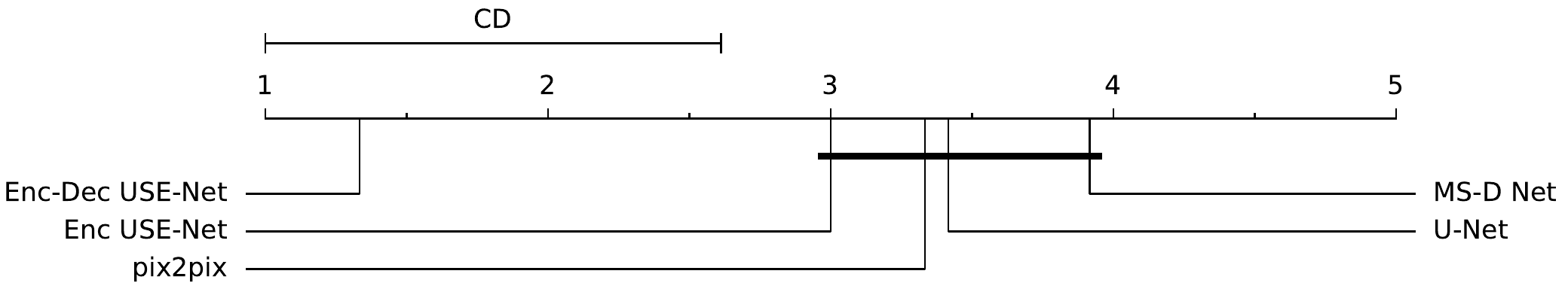}\label{sfig:BonferroniCG_triple}} \\
    \subfloat[][Peripheral Zone]
    {\includegraphics[width=\textwidth]{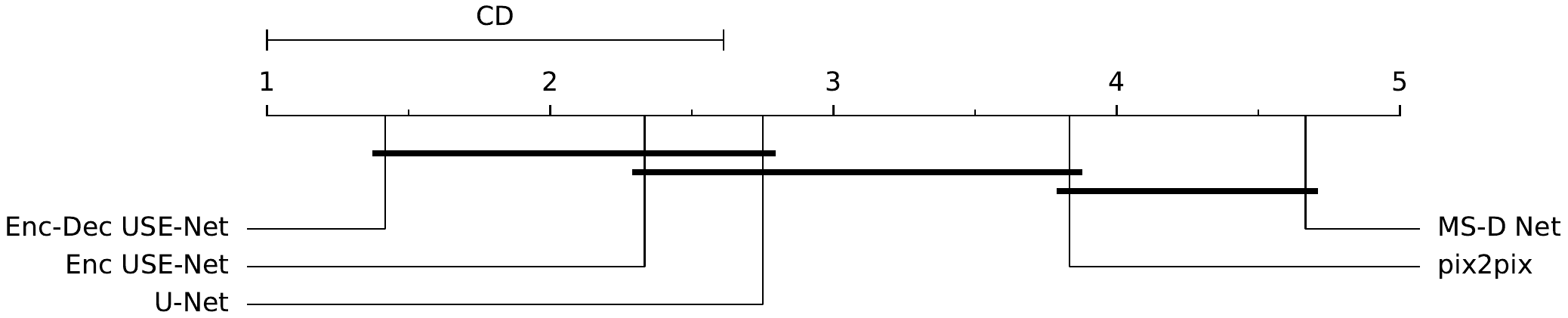}\label{sfig:BonferroniPZ_triple}} \\ 
  \caption{Critical Difference (CD) diagram comparing the \textit{DSC} values achieved by all the investigated CNN-based architectures using the Bonferroni-Dunn's \textit{post hoc} test~\cite{demvsar2006} with $95\%$ confidence level for the three-dataset training conditions. Bold lines indicate groups of methods whose performance difference was not statistically significant.}
  \label{fig:Bonferroni_three}
\end{figure}

We executed the Friedman's test to quantitatively investigate any statistical performance differences among the tested approaches.
Regarding the three-dataset condition: $p=0.0009$ and $p=1.3 \cdot 10^{-6}$ for the CG and PZ, respectively.
Considering all training/testing combinations: $p=0.01$ and $p=1.7 \cdot 10^{-10}$ for the CG and PZ, respectively.
Since the $p$-values allowed us to reject the null hypothesis, we performed the Bonferroni-Dunn's \textit{post hoc} test for both the three-dataset condition and all training/testing combinations~\cite{demvsar2006}.
In order to visualize the achieved results, example images segmented by each method are compared in Fig. \ref{fig:segResD} under the three-dataset training condition.
The critical difference diagram (Fig. \ref{fig:Bonferroni_three}) using the Bonferroni-Dunn's \textit{post hoc} test also confirms this trend, considering \textit{DSC} values for every round of the $4$-fold cross-validation.

\begin{figure}[ht!]
\centering
	\subfloat[][Central Gland]
    {\includegraphics[width=\textwidth]{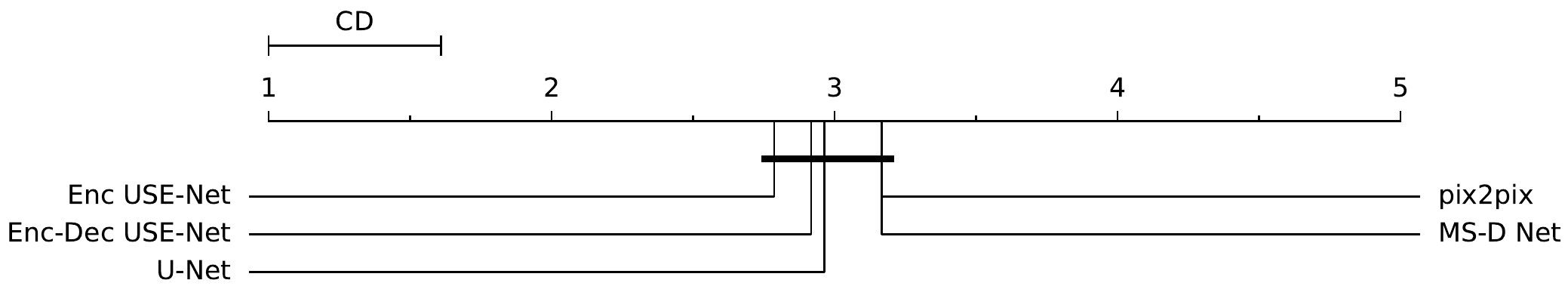}\label{sfig:BonferroniCG_all}} \\
    \subfloat[][Peripheral Zone]
    {\includegraphics[width=\textwidth]{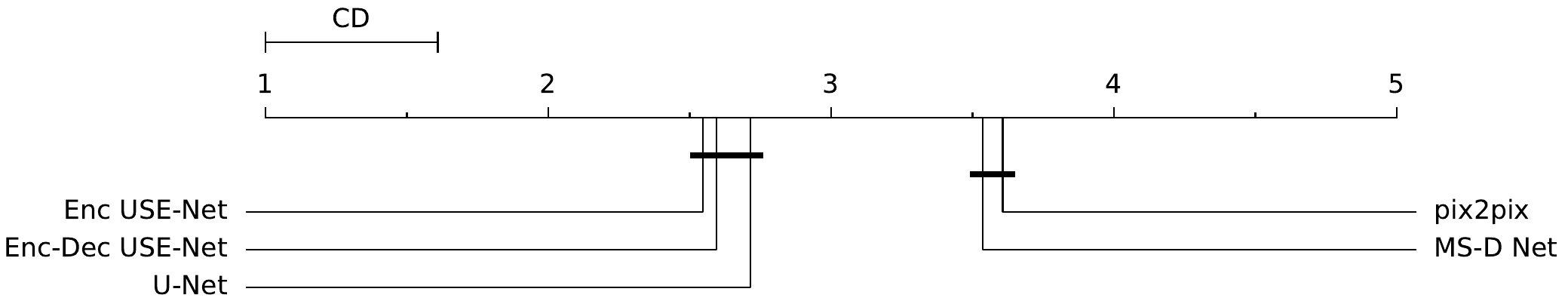}\label{sfig:BonferroniPZ_all}} \\ 
  \caption{Critical Difference (CD) diagram comparing the \textit{DSC} values achieved by all the investigated CNN-based architectures using the Bonferroni-Dunn's \textit{post hoc} test~\cite{demvsar2006} with $95\%$ confidence level considering all training/testing combinations. Bold lines indicate groups of methods whose performance difference was not statistically significant.}
  \label{fig:Bonferroni_all}
\end{figure}

However, as shown in Fig. \ref{fig:Bonferroni_all}, Enc-Dec USE-Net shows less powerful cross-dataset generalization when trained and tested on different datasets, achieving slightly lower average performance than Enc USE-Net (considering all training/testing combinations).
This implies that the SE blocks' adaptive feature recalibration---boosting informative features and suppressing weak ones---provides excellent intra-dataset generalization in the case of testing performed on multiple datasets used during training (i.e., when training samples from every testing dataset are fed to the model).

On the contrary, pix2pix achieves good generalization when trained and tested on different datasets, especially under mixed-dataset training conditions, thanks to its internal generative model.
MS-D Net generally works better in single dataset scenarios, using a limited amount of training samples, according to \cite{pelt2017}.
The unsupervised continuous max-flow model achieves comparable results to the supervised ones only when trained and tested on different datasets.
However, this semi-automatic approach is outperformed by the supervised methods when trained and tested on the same datasets, as it underestimates $\mathcal{R}_{CG}$.

The results also reveal that training on multi-institutional datasets generally outperforms training on each dataset during testing on any dataset/zone, realizing both intra-/cross-dataset generalization.
For instance, training on datasets $\#1$ and $\#2$ generally outperforms training on dataset $\#1$ during testing on all datasets $\#1$, $\#2$, and $\#3$, without losing accuracy. 

Therefore, training schemes with mixed MRI datasets can achieve reliable and excellent performance, potentially useful for other clinical applications.
Comparing the CG and PZ segmentation, the results on the CG are generally more accurate, except when trained and tested on dataset $\#1$; this could be due to intra- and cross-scanner generalization, since dataset $\#1$'s scanner is different from those of datasets $\#2$ and $\#3$.

The trend characterizing the best \textit{DSC} accuracy performance, especially in the case of three-dataset training/testing conditions, is reflected by both the \textit{SEN} and \textit{SPC} values (Tables S1 and S2).
As shown in Tables S3 and S4, the achieved spatial distance-based indices are consistent with overlap-based metrics.
Hence, Enc-Dec USE-Net obtained high performance also in terms of difference between the automated and the manual boundaries.

Considering more permutations in the random partitioning and running multiple $4$-fold cross-validation instances may increase the robustness of the results, by evaluating the combination of the multiple executions.
However, with particular reference to the three-dataset training/testing condition, where the feature recalibration can effectively capture the dataset characteristics with the most available samples, the Bonferroni-Dunn’s \textit{post hoc} test showed significant differences in the multiple comparisons among the competing architectures (Fig.~\ref{fig:Bonferroni_three}).
On the contrary, no significant statistical difference was detected when considering all training/testing conditions (Fig.~\ref{fig:Bonferroni_all}).
The achieved results suggest that cross-validation with a single random permutation is methodologically sound.
In addition, we can state that the patterns arising from the $4$-fold cross-validation experiments are not just by chance or biased by the increased training samples, so USE-Net significantly outperforms the other techniques.

To conclude, the comparison of U-Net and USE-Nets shows the individual contribution of SE blocks under each of the $21$ dataset combinations.
Interestingly, USE-Net is not always superior on one- or two-dataset cases, but consistently outperforms U-Net on three-dataset training/testing.
This arises from USE-Net's higher number of parameters than U-Net, generally requiring more samples for proper tuning.

\section{Discussion and Conclusions}
\label{sec:Conclusions}

The novel CNN architecture introduced in this work, Enc-Dec USE-Net, achieved accurate prostate zonal segmentation results when trained on the union of the available datasets in the case of multi-institutional studies---significantly outperforming the competitor CNN-based architectures, thanks to the integration of SE blocks~\cite{hu2017} into U-Net~\cite{ronneberger2015}.
This also derives from the presented cross-dataset generalization approach among three prostate MRI datasets, collected by three different institutions, aiming at segmenting $\mathcal{R}_{CG}$ and $\mathcal{R}_{PZ}$; Enc-Dec USE-Net's segmentation performance considerably improved when trained on multiple datasets with respect to individual training conditions.
Since the training on multi-institutional datasets analyzed in this work achieved good intra-/cross-dataset generalization, CNNs could be trained on multiple datasets with different devices/protocols to obtain better outcomes in clinically feasible applications. 
Moreover, our research also implies that state-of-the-art CNN architectures properly combined with innovative concepts, such as feature recalibration provided by the SE blocks~\cite{hu2017}, allow for excellent intra-dataset generalization when tested on samples coming from the datasets used for the training phase.
Therefore, we may argue that multi-dataset training and SE blocks represent not just individual options but mutually indispensable strategies to draw out each other's full potential.
In conclusion, such adaptive mechanisms may be a valuable solution in medical imaging applications involving multi-institutional settings.

As future developments, we will refine the output images considering the 3D spatial information among the prostate MR slices.
Finally, for better cross-dataset generalization, we plan to use domain adaptation \textit{via} transfer learning by maximizing the distribution similarity \cite{vanOpbroek2015}.
In this context, Generative Adversarial Networks (GANs)~\cite{goodfellow2014,han2018} and Variational Auto-Encoders (VAEs)~\cite{kingma2013} represent useful solutions.

\section*{Acknowledgment}
This work was partially supported by the Graduate Program for Social ICT Global Creative Leaders of The University of Tokyo by JSPS.

We thank the Cannizzaro Hospital, Catania, Italy, for providing one of the imaging datasets analyzed in this study.

\section*{References}
\bibliographystyle{elsarticle-num}
\bibliography{biblio.bib}







\end{document}